\documentclass[times,3p,preprint]{elsarticle}
\journal{journal}
\usepackage{graphicx}     
\usepackage{txfonts}

\usepackage[utf8]{inputenc}
\usepackage[english]{babel}

\usepackage{times,mathtools,amsfonts}
\usepackage{helvet,courier}
\usepackage{txfonts}
\usepackage{amsthm}
\usepackage{amssymb}
 
\theoremstyle{definition}
\newtheorem{definition}{Definition}[section]

\usepackage{natbib}
\usepackage{graphicx}
\usepackage{amsfonts}
\usepackage{amsmath}
\usepackage{hyperref}
\usepackage{tikz}
\usepackage{mdsymbol}
\usetikzlibrary{shapes,arrows}

\newcommand{\ci}{\mathrel{\text{\scalebox{1.07}{$\perp\mkern-10mu\perp$}}}}

\begin{document}
	\begin{frontmatter}

	\title{Fairness in Machine Learning with Tractable Models}
		


				%
				\cortext[cor]{Corresponding author. Vaishak Belle was supported by a Royal Society University Research Fellowship.}
				 \address[edinburgh]{School of Informatics, University of Edinburgh,
				Edinburgh, UK.}
				 \address[turing]{Alan Turing Institute,
				London, UK.}

 \author[edinburgh]{Michael Varley}

				\author[edinburgh,turing]{Vaishak Belle\corref{cor}}
				\ead{vaishak@ed.ac.uk}


%
%
%
%
%
%
%
%
%
%
%

\begin{abstract}
	
	Machine Learning techniques have become pervasive across a range of different applications, and are now widely used in areas as disparate as recidivism prediction, consumer credit-risk analysis and insurance pricing. The prevalence of machine learning techniques has raised concerns about the potential for learned algorithms to become biased against certain groups. Many definitions have been proposed in the literature, but the fundamental task of reasoning about probabilistic events is a challenging one, owing to the intractability of inference. 

    The  focus of this paper is taking steps towards the application of tractable probabilistic models to  fairness in machine learning. Tractable probabilistic models have recently emerged that guarantee that conditional marginal can be computed in time linear in the size of the model. 
		%
	In particular, we show  that sum product networks (SPNs) enable an effective technique for determining the statistical relationships between protected attributes and other training variables. We will also motivate the concept of ``fairness through percentile equivalence'', a new definition predicated on the notion that individuals at the same percentile of their respective distributions should be treated equivalently, and this prevents unfair penalisation of those individuals who lie at the extremities of their respective distributions.

	  We compare the efficacy of this pre-processing technique with an alternative approach that assumes an additive contribution (used in  \citep{Counterfactual_Fairness}). It was found that when these two approaches were compared on a data set containing the results of law school applicants \citep{lsac_information}, the percentile equivalence method reduced the average underestimation in the exam score of ethnic minority applicants black applicants at the bottom end of their conditional distribution by about a fifth. We conclude by outlining potential improvements to our existing methodology and suggest opportunities for further work in this field.

\end{abstract}
%
%
\end{frontmatter}

\section{Introduction}

Machine Learning techniques have become pervasive across a range of different applications, and are now widely used in areas as disparate as recidivism prediction, consumer credit-risk analysis and insurance pricing \citep{Recidivism_study, ML_for_credit}. The prevalence of machine learning techniques has raised concerns about the potential for learned algorithms to become biased against certain groups. This issue is of particular concern in cases when algorithms are used to make decisions that could have far-reaching consequences for individuals (for example in recidivism prediction) \citep{Recidivism_study, ProPublica}. Attributes which the algorithm should be ``fair'' with respect to are typically referred to as \textbf{protected} attributes, and this terminology is used throughout the paper. There are multiple different potential fields that might qualify as protected attributes in a given situation, including ethnicity, sex, age, nationality and marital status \citep{Learning_Fair_Reps}. Ideally, such attributes should not affect any prediction made by a truly ``fair'' machine learning algorithm. However, even in cases where it is clear which attributes should be protected, there are multiple (and often mutually exclusive) definitions of what it means for an algorithm to be unbiased with respect to these attributes, and there remains a stark lack of agreement on this subject within the academic community \citep{Fairness_through_awareness, Counterfactual_Fairness, Disparate_Impact_&_Historical}.

\renewcommand{\newblock}{}
\newblock

Even if algorithms are not provided with information about the protected attribute directly, there is still the potential for algorithms to discriminate on the basis of \textbf{proxy} variables, which may contain information about the protected attribute \citep{Causal_Reasoning}. For example, the name of an individual may not be regarded as a protected attribute, but it will likely contain information about an individual's ethnicity or sex. Unfair algorithmic biases can also be introduced if an algorithm is trained on historical data, particularly in cases where the training labels were allocated on the basis of human discretion \citep{Disparate_Impact_&_Historical, tick_cross_paper}. In such a scenario, the algorithm can inherit the historical biases of those responsible for allocating the labels, by actively discriminating against attributes prevalent in a particular group \citep{Disparate_Impact_&_Historical, Learning_Fair_Reps}. Similarly, an algorithm may be unfair if (at test time) any input attribute was decided by a human (e.g. risk-score of reoffence based on expert's opinion). In such a scenario, the outcome of the prediction will inevitably be influenced by the biases of the individual allocating the score if not adjusted accordingly.

In recent years, the topic of fairness has become an increasingly important  issue within the field of machine learning, both in academic circles \citep{Counterfactual_Fairness, Causal_Reasoning, Equality_of_opportunity}, and more recently in the public and political domains \citep{ProPublica, White_House_Report}. However, even amid pressing concerns that algorithms currently in use may exhibit racial biases, there remains a lack of agreement about how to effectively implement fair machine learning algorithms within the academic community.

\subsection{Motivation}

This paper is primarily concerned with the application of tractable inference techniques to problems in fair machine learning. This is principally motivated by the observation that the fundamental challenge of reasoning about probabilistic events, as one would to contextualize fairness, is a challenging one, owing to the intractability of inference \cite{DBLP:journals/jair/BacchusDP09}. 

Of course, the intractability of probabilistic inference is a challenge even outside fairness concerns. In that regard, probabilistic tractable models have  emerged that guarantee that conditional marginal can be computed in time linear in the size of the model, which is often learned from data. While initially limited to low tree-width models \cite{bach2002thin}, recent tractable models such sum product networks (SPNs)  \cite{SPN,SPN_structure_learning} and probabilistic sentential decision diagrams (PSDDs) \cite{kisa2014probabilistic} are derived from arithmetic circuits  \cite{darwiche2002logical,choi2017relaxing} that capture local structure and determinism.  

In this work, we focus on SPNs; however,  related data structures may also be considered, e.g.,  \cite{liang2017learning}. 
We hypothesise that by learning the structure of a SPN from data, we can establish which training attributes exhibit strong statistical dependency on a protected attribute \citep{SPN_structure_learning}. Furthermore, we suggest that by performing inference in the resulting SPN, we can remove the contribution of the protected attribute to the other available variables, and thus train a fair classifier on the adjusted data. We elected to use this particular technique because, unlike many other graphical models, the computational cost of inference queries scales linearly with the total number of edges in the SPN architecture \citep{SPN_structure_learning}. We argue that this tractability property makes SPNs  appropriate for the creation of fair yet {robust} classifiers trained on large data sets.


We then address some issues with our initial techniques and explore scenarios in which the contribution of the protected attribute to the training variables isn't an additive constant. To address this issue, we introduce the notion of ``fairness through percentile equivalence'', which is presented in the final section of this report. We believe that this criterion offers a solution to some of the problems associated with ensuring that individuals at the extremities of their conditional distributions are treated fairly, and demonstrate how a simple additive model could result in unfair treatment of certain individuals even if it improves statistical parity for the group as a whole.\footnote{We developed this idea independently as a method of ensuring that pre-processing techniques don't lead to unfair treatment of outlying individuals. However, at the time of writing of this paper, 
it was brought to our attention recently that similar ideas have  been explored  by Feldman et. al. \citep{Feldman}. In that regard, it would be worthwhile to consider a comprehensive empirical comparison to \citep{Feldman}, which we leave for future work.} 

{We reiterate that 
the primary thrust of this work is in understanding how tractable learning and fairness can come together, and in this regard, we are deliberately expository in some parts. While we do contribute  technical insights in adapting a standard fairness criteria to tractable learning and investigating percentile equivalence (as discussed above), our broader goal is to make the fairness literature accessible to tractable learning researchers.} 


\subsection{Problem Context}

Given the potential ethical and legal ramifications of deploying algorithms which exhibit bias, the motivation for building a general-purpose framework for fair learning in large data sets is immediate. An internal White House report on the impact of big data produced for the Obama administration in 2014 gave special mention to the topic of bias in machine learning algorithms and identified it as one of the main areas of concern for policy makers \citep{White_House_Report}. At present, there is no particular fairness criterion which is universally agreed upon, though several have been proposed in recent years \citep{Learning_Fair_Reps,Fairness_through_awareness,Counterfactual_Fairness, Equality_of_opportunity} and we will discuss the merits and shortcomings of several of these in detail in the literature review.

\newblock

One well-publicised recent episode serves to highlight the necessity for appropriate application of fairness criteria to machine learning models when testing for bias. Pro-Publica, a US-based entity specialising in not-for-profit journalism, published an article suggesting that an algorithm widely used to predict the probability of re-offense in criminals was biased against black offenders \citep{ProPublica}. The article raised concerns about the fairness and efficacy of the Correctional Offender Management Profiling for Alternative Sanctions (or COMPAS) algorithm, which is widely used in the US justice system \citep{ProPublica}. Their article received criticism from both members of the academic community \citep{ProPublica_Criticism_Academics} and Northpointe, the company who created the algorithm \cite{Northpointe_refute}. Their primary complaint concerned the metric used by ProPublica to measure discrimination; the original concerns about racial bias were based mainly on the discrepancy in false positive and false negative rates between black and white offenders \cite{ProPublica_Criticism_Academics, ProPublica}. This analysis was critiqued in part because the authors of \citep{ProPublica} failed to appreciate that the outcome of the algorithm was not a prediction of future behaviour per se, but actually a risk allocation \citep{ProPublica_Criticism_Academics}. ProPublica defined a false-positive as any individual considered ``high-risk'' who didn't re-offend \citep{ProPublica}, whereas in reality the risk categories were simply an indication of reoffence probability. Moreover, Northpointe and others illustrated that the algorithm satisfied a calibration criterion \citep{Northpointe_refute, Recidivism_study}, which they regarded as a more appropriate measure of fairness in this particular instance. This calibration requirement states that if an algorithmic process maps a series of inputs $X$ to a score $s$, such that:
\begin{equation}
    f:X\rightarrow{}s,
\end{equation}
then the probability of re-offence (denoted $r=1$) of individuals at every value of $s$ is independent of the value of the protected attribute $a_p$ (which in this case is race) \citep{Recidivism_study}, i.e.
\begin{equation}
    P(r=1|a_p,s)=P(r=1|s) \hspace{2mm} \forall \hspace{1mm} s,a_p.
\end{equation}

\newblock

To be explicit, Northpointe claim that their algorithm is fair, since individuals allocated the same risk level are equally likely to re-offend regardless of ethnicity \citep{Northpointe_refute}. They attribute the higher ``false positive'' rate in black offenders to the fact that, statistically, a greater fraction of the underlying black population went on to re-offend, and therefore a larger proportion will be allocated a higher risk score if the calibration criterion is to be satisfied. If (say) 50\% of individuals considered ``high-risk'' go on to re-offend, regardless of their ethnicity, then it is statistically guaranteed that the population with the higher re-offence rate will see a greater proportion of the TOTAL population be allocated to the ``high-risk'' category without going on to re-offend \citep{Recidivism_study}. In reality, it's impossible to satisfy the calibration criterion and achieve identical false negative rates unless the recidivism rate were identical in the underlying populations \citep{Northpointe_refute, ProPublica_Criticism_Academics,Recidivism_study}.

\newblock

This episode provides an example of a situation where there is the potential for injustices to arise as a consequence of bias exhibited by machine learning algorithms on the basis of sensitive factors. Ideally, such factors should be outside the purview of the algorithm's decision making process. Moreover, it is apparent that many criteria used to determine fairness are mutually incompatible, and that caution should be used when selecting an appropriate fairness criterion for a specific situation. This can lead to significant discrepancies in the interpretation of the same set of results. 

\newblock

Despite the concerns listed above and some recent negative coverage in the media, it is important to emphasize that automating classification processes actually offers significant opportunity for improvement. Algorithmic classification takes the decision process out of the hands of humans, who might exhibit significant bias in their decision making process. A recent report published by the National Bureau of Economic research in the US found that the application of machine learning algorithms to deciding the eligibility for pre-trial release of defendants led to lower rates of incarceration in black and Hispanic individuals, without any increase in criminal activity attributed to individuals on bail \citep{NBER_pre_trial_release}. In this case, the use of the technology compared favourably to human judgment \citep{NBER_pre_trial_release}. Moreover, despite the lack of universal agreement over what constitutes a ``fair'' classifier, the decisions made by algorithmic processes can still be assessed through several different measures of bias. This gives those responsible for maintaining the algorithm a quantitative understanding of how the classifier is treating different demographic groups. Fair machine learning should therefore be regarded as a good opportunity to negate human biases and improve prediction accuracy in a fair manner.

\subsection{Paper Outline}

This paper explores the application of tractable  learning techniques (specifically SPNs) to problems in fair machine learning. 

%

We start by providing an introduction to the topic of fairness in machine learning. A number of different fairness criteria are outlined, and some of the advantages and shortcomings of these criteria are discussed. We also discuss certain previous techniques for training fair classifiers, outlining the advantages and disadvantages of pre-processing and intra-processing techniques. We then review SPNs, which can be learnt from data in an unsupervised manner \citep{SPN_structure_learning}. 

%

Our contribution involves the construction of a new procedure for pre-processing data to ensure fair training. We start by introducing a slight modification to the original code used in \citep{SPN_structure_learning} which allows us to identify subsets of mutually independent variables within a training set. This allows us to identify a collection of `safe' variables, which can be used to train a fair model since they don't depend on any protected attributes. We then move on to consider how an SPN can be used to predict the contribution made by the protected attribute to other training variables, and we suggest a method to compensate for this in the case of both numeric-valued and qualitative variables.

\newblock

Our results illustrate that this pre-processing technique leads to a significant reduction in the disparate treatment of males and females within the German Credit data set \citep{German_Credit_Source} relative to a ``fairness through unawareness'' approach. Moreover, the effect on the classification accuracy of the two models is very limited. This initial result is encouraging, as alternative approaches have sometimes led to more substantial reductions in classification accuracy when attempting to compensate for fair treatment \citep{Learning_Fair_Reps}.

\newblock

Following these results, we extend our ideas still further, and introduce the notion of ``fairness through percentile equivalence''. We demonstrate that individuals at the far end of a spectrum can be treated unfairly by systems which impose an additive adjustment to all members of a particular group based on their protected attribute, and introduce a more nuanced technique to correct for this. We found that, on a simple data set of numerical attributes, our technique led to a minor improvement in the fair treatment of individuals at the bottom end of the conditional distribution. Finally, we suggest certain alterations to the SPN architecture and our experiments which could be built upon in future work, and identify opportunities for comparison with other techniques.

\renewcommand{\newpage}{}
\newpage

\section{Background and Overview}

\subsection{Fairness}
Before discussing the techniques used in this paper, we begin with an overview of the recent literature in the field of fair machine learning. There have been several papers published in this field in recent years, yet despite this there is still no universally agreed definition of what it means for an algorithm to be fair \citep{Causal_Reasoning,Counterfactual_Fairness,Learning_Fair_Reps,Equality_of_opportunity}. In truth, the most appropriate criterion probably differs on a case-by-case basis, and depends on whether or not the labels being used by the algorithm may be subject to human biases \citep{Disparate_Impact_&_Historical}. We commence this section by describing some of the better known criteria and discuss their advantages and shortcomings. Throughout this section, we denote the protected attribute with respect to which we wish to be impartial by $a_p$. A general set of training variables is denoted $X$ and the true labels associated with each individual are denoted $y$. Following the notation used in \citep{Counterfactual_Fairness}, we denote the prediction of each individual by $\hat{y}$.

\subsubsection{Fairness Through Unawareness}
Fairness through unawareness is the simplest definition of fairness; as its name suggests, under this definition an algorithm is “fair” if it is unaware of the protected attribute $a_p$ of a particular individual when making a prediction \citep{Counterfactual_Fairness}. 
\begin{definition}{Fairness Through Unawareness}\newline
    For some set of attributes $X$ any mapping
    \begin{equation}
        f:X\xrightarrow{}\hat{y}
    \end{equation}
    where $a_p \not \in X$ satisfies fairness through unawareness \citep{Counterfactual_Fairness}.
\end{definition}

This prevents the algorithm learning direct bias on the basis of the protected attribute, but doesn’t prevent indirect bias, which the algorithm can learn by exploiting the relationship between other training variables and the protected attribute \citep{Discrimination_Aware_DM, Equality_of_opportunity}. Moreover, if any of the training attributes are allocated by humans there is the potential for bias to be introduced. Nonetheless, it is extremely simple to guarantee that an algorithm is fair in this sense. Throughout this paper, we use fairness through unawareness as a comparative baseline technique, and attempt to illustrate how backing out the contribution of the protected attribute to the remaining training variables can reduce bias.

\subsubsection{Statistical Measures of Fairness}
Rather than defining fairness in terms of the scope of the training data, much of the existing literature instead assesses whether an algorithm is fair on the basis of a number of statistical criteria that depend on the predictions made by the algorithm \citep{Equality_of_opportunity, Counterfactual_Fairness, Learning_Fair_Reps}. One widely used and simple criterion is demographic parity. In the case that both the predicted outcome and protected attribute $a_p$ are both binary variables, a classifier is said to satisfy predictive parity if $P(y=1|a_p=1) = P(y=1|a_p=0)$ \citep{Equality_of_opportunity}. By this definition, a classifier is considered fair if it is equally likely to make the prediction $\hat{y}=1$ regardless of the value of the protected attribute $a_p$.  

\newblock

One problem with this definition is that (unless the instance rate of $y=1$ happens to be the same in both the $a_p=0$ group and $a_p=1$ group), the classifier cannot achieve 100\% classification accuracy and satisfy the fairness criterion simultaneously \citep{Equality_of_opportunity}. Also, there are scenarios where this definition is completely inappropriate because the instance rate of $y=1$ differs so starkly between different demographic groups. For example, consider the case of recidivism prediction. On average, men are far more likely to be arrested than women, and women comprise less than 10\% of the incarcerated population in the United Kingdom \citep{UK_criminal_justice}. A hypothetical system that insisted on predictive parity between sexes for prediction of future criminality would unfairly penalise women, and an alternative criterion would be more applicable in this case \citep{Equality_of_opportunity}. Finally, there are also concerns that statistical parity measures fail to account for fair treatment of individuals \citep{Fairness_through_awareness}, a topic we will explore more when considering the shortcomings of fairness through intra-processing. Nonetheless it is widely used as a target and is often regarded as the most appropriate statistical definition when an algorithm is trained on historical data \citep{tick_cross_paper,Learning_Fair_Reps}.

\newblock

A modification of demographic parity is ``equality of opportunity''. By this definition, a classifier is considered fair if, among those individuals who meet the positive criterion, the instance rate of correct prediction is identical, regardless of the value of the protected attribute \citep{Equality_of_opportunity}. Mathematically, this condition can be expressed as follows \citep{Equality_of_opportunity}:
\begin{equation}
    P(y=1|a_p=a,\hat{y}=1)=P(y=1|a_p=a',\hat{y}=1) \hspace{2mm} \forall \hspace{1mm} a,a'
\end{equation}
The authors of \citep{Equality_of_opportunity} point out that a classifier can simultaneously satisfy equality of opportunity and achieve perfect prediction whereby $\hat{y}=y$ in all cases. The equality of opportunity criterion might well be better applied in instances where there is a known underlying discrepancy in positive outcomes between two different groups, and this discrepancy is regarded as permissible.

\newblock

However, concerns have been raised about the applicability of this metric to data sets where the training data already contains historical biases, and either the training data or labels have had human input \citep{Counterfactual_Fairness}. To make this point explicit, consider a scenario where the labels for a training set were allocated by humans. In such a situation, it is possible that the lower rate of positive ($y=1$) outcomes in a particular group are attributable to the biases of the individuals responsible for allocating the labels. Therefore, demanding that a classifier satisfies equality of opportunity ensures that pre-existing historical biases will continue to perpetuate. 

\subsubsection{Fairness and the Individual}

Another problem with statistical measures is that, provided that the criterion is satisfied, an algorithm will be judged to be fair regardless of the impact on individuals. For example, consider a classifier which predicts debt default risk for individuals. Even if a `fair' algorithm incorrectly classifies certain low-risk individuals as high risk, it can still be deemed `fair' according to the statistical definitions, despite the fact that the incorrectly classified individuals may regard themselves as having been treated unfairly. Therefore, these statistical metrics may be regarded as incomplete, even if they are useful in certain circumstances \citep{Fairness_through_awareness}.

\newblock

In view of these limitations, various works have introduced fairness metrics which aim to ensure that individuals are treated fairly, rather than simply considering the statistical impact on the population as a whole \citep{Fairness_through_awareness, Counterfactual_Fairness}. Counterfactual fairness, for example,  was proposed as a fairness criterion in  \citep{Counterfactual_Fairness}. The fundamental principle behind this definition of fairness is that the outcome of the algorithm's prediction should not be altered if different individuals within the sample training set were allocated different values for their protected attributes \citep{Counterfactual_Fairness}. This criterion is written  in the following form:
\begin{equation}
    P(\hat{y}_{A_p \leftarrow a_p}|A=a, X=x) = P(\hat{y}_{A_p \leftarrow a_p'}|A=a, X=x) \hspace{1mm}\forall y,a'.
\end{equation}

\newblock

The notation $\hat{y} \leftarrow _{A_p \leftarrow a_p}$ is understood as ``the value of $\hat{y}$ if $A_p$ had taken the value $a_p$'' \citep{Counterfactual_Fairness}. This is clearly a more difficult criterion to demonstrate empirically than the statistical attributes mentioned above, since we only have knowledge about the ``true'' outcomes and data, and don't a priori have access to the ``counterfactual'' data that we would have been presented with if different individuals within the sample had different protected attributes \citep{Counterfactual_Fairness}. The necessity to infer counterfactual values and the difficulties associated with the backing out the causal contribution of protected attributes are often regarded as the major drawbacks of such an approach \citep{Causal_Reasoning}. 

\newblock

In order to compute such counterfactual values, the authors postulate the existence of unseen background variables $U_i$, which are not caused by any visible variables including the protected attribute. Thus, they regard any visible variable $V_i$ as being a stochastic function of its parents (both observed and unobserved) in a directed graphical model (DGM)  $V_{pa_i} \cup U_{pa_i}$ \citep{Counterfactual_Fairness}. If some subset of all visible variables $V$ are non-descendants of the protected attribute in the graphical model, then it should be apparent that any model trained on just that subset is counterfactually fair with respect to the protected attribute \citep{Counterfactual_Fairness}.

\newblock

\begin{figure}[h!]
\centering
\begin{tikzpicture}[->,>=stealth',shorten >=1pt,auto,node distance=3cm,
                    thick,main node/.style={circle,draw,font=\sffamily\Large\bfseries}]

  \node[main node] (s1)  at (-2, 0)  {$U_k$};
  \node[main node] (s2)  at (-6, 0) {$V_j$};
  \node[main node] (s3)  at (-2, -4) {$y$};
  
  \node[main node] (r1) at  (2, 0)   {$a_p$};
  \node[main node] (r3)  at (-2, -2) {$V_k$};

  \path[->]
   (r1) edge (r3)
   (s1) edge (r3)
   (s2) edge (r3)
   (r3) edge (s3)
   (s2) edge (s3);
\end{tikzpicture}
\caption{General structure outlined for causal directed graphical model in \citep{Counterfactual_Fairness}.}
\label{fig:dirgraph_1}
\end{figure}
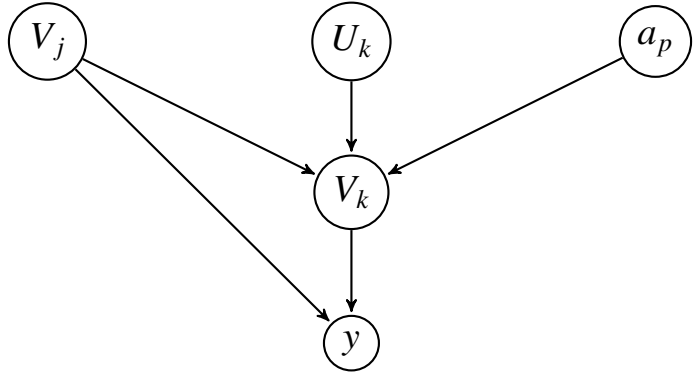

However, some of the $V_i$ in the DGM will be descendants of the protected attribute, and therefore any model trained using the unadjusted $V_i$ may not be counterfactually fair, as an adjustment in the value of the protected attributes would also result in an adjustment to the $V_i$ inputted to the model, causing the parameters of the final predicted model to change and the values of the output $\hat{y}$ to be altered. Finally, in  \citep{Counterfactual_Fairness}, the authors propose a mechanism for training on the basis of these attributes in a fair way. They postulate that there exists an unobserved $U_i$ for all observed $V_i$, and that the $V_i$ is a linear combination of the contribution of the protected attribute, $c(a_p)$ and the latent variable:
\begin{equation}
    V_i = U_i + c(a_p).
\end{equation}
One can then train a simple model using only the protected attribute to predict the value of each $V_i$, and regard the outcome of that model as the contribution of the protected variable to the visible $V_i$. Subtracting this leaves a residual, which can be interpreted as the value of the hidden variable $U_i$. One plausible structure for a causal model with only two visibles $V_i$ and a single protected attribute $a_p$ is shown in figure \ref{fig:dirgraph_1}.

By construction, the $U_i$ are not causally dependent on the protected attribute $a_p$, and so we can safely train a model using the $U_i$ derived from the $V_i$. Ultimately, this is a data pre-processing strategy which notionally removes the contribution of the protected attribute to each variable. In figure \ref{fig:dirgraph_1}, a counterfactually fair model may be trained using the values of $U_k$ and $V_j$ \citep{Counterfactual_Fairness}. 

\newblock

There are several advantages to this approach, particularly from a theoretical standpoint. Firstly, if we define a fair algorithm as one where the protected attribute plays no role in the algorithm's outcome, then counterfactual fairness is arguably an optimal definition, as the outcome must be the same regardless of the values of the protected attributes in the data set. Secondly, as a definition it holds at an individual level, which negates the problem encountered by the statistical measures - namely that certain individuals may receive unfair punishment in order to artificially satisfy the statistical criterion \citep{Fairness_through_awareness}.

\newblock

However, the actual implementation of a counterfactually fair algorithm presents us with several difficulties. As already noted, computing counterfactual values requires the use of strong assumptions about the model and the introduction of latents \citep{Counterfactual_Fairness, Causal_Reasoning}. A more practical problem is that, whilst the notion of counterfactual values is clearly applicable to continuous real-valued variables, it is unclear how the notion of a counterfactual value could be extended to the types of discrete categorical variables typically found in real world data. In the continuous case, a simple statistical analysis can lead to the conclusion that having protected value attribute $a_p=0$ contributes $x_0$ to continuous variable $V_i\in \mathbb{R}$, and that this can subsequently be corrected for by subtracting $x_0$. However, for discrete categorical variables, which take on of a finite number of values (e.g. employed - yes/no, or ``region of residence'' - A,B,C,D), there is no obvious method by which the ``contribution'' of the protected attribute to these criteria can be compensated for to create a counterfactual value. There appears to be little mention of this particular issue in the literature, but it certainly led to conceptual problems when we attempted to train a counterfactually fair model using SPNs as discussed in the sequel. 

\newblock

Furthermore, the assumption about the contribution of the protected criteria is relatively simplistic. The authors assume a constant additive contribution from the protected attribute to the variable \citep{Counterfactual_Fairness}.  However, we argue later that to truly remove the contribution, one should attempt to match the probability distributions over the entire domain of the attribute. 

\newblock

Finally, despite assuming the existence of a causal directed graphical model, the authors do not suggest how such a model might be constructed. This is not quite as problematic in their illustrative example, where the entire data set comprises just two continuous-valued training variables, a target variable and a single protected attribute \citep{Counterfactual_Fairness}. However in a real world data set, which may potentially comprise dozens of attributes, there is no method for establishing independence between different attributes. Nonetheless, this definition successfully avoids the shortcomings of the statistical measures mentioned earlier, and much of this paper's focus is on attempting to address some of these problems.

\subsubsection{Causal Reasoning}
More recently, certain papers have shifted the focus of fair machine learning from defining fairness criteria to assessing the causal effect of the protected attribute on the outcome of the model \citep{Causal_Reasoning}. As with counterfactual fairness, implementations of this idea are dependent on having access to a causal model, which are difficult to infer from data \citep{Causal_Reasoning}. As in the counterfactual case, the authors of \citep{Causal_Reasoning} advocate training on adjusted data of the form:
\begin{equation}
    \tilde{X} = X - \mathbb{E}(X|P)
\end{equation}
where $P$ is some proxy containing information about the protected. Thus, in this sense, the contribution of the protected is ``backed-out'' before the model is trained. These adjusted variables are in some sense analogous to the latent variables calculated in \citep{Counterfactual_Fairness}, but now they are removed as removing an arrow in the causal model from the proxy to the variable, rather than introducing a new latent that's independent of the protected attribute. Again, in many cases the authors assume knowledge about causal relationships between variables, which may not be obvious a priori.

\subsubsection{Implementing Fairness}

We provide here a brief overview of attempts to implement some of the definitions of fairness described above. The approaches can be broadly partitioned into two categories: `intra-processing' regularisation techniques which aim to minimise a measure of fairness whilst training is underway (e.g. \citep{Learning_Fair_Reps, Disparate_Impact_&_Historical, tick_cross_paper}), and `pre-processing' techniques, where the data is modified in the hope of removing any dependence on the protected attribute prior to training \citep{Counterfactual_Fairness, Calders_first_latent_attempt, Causal_Reasoning}. Sometimes, these pre-processing techniques have been as simple as modifying training labels until the training set achieves demographic parity \citep{kNN_point_removal, Calders_first_latent_attempt}. Similar ideas to those put forward in \citep{Counterfactual_Fairness} were also explored in earlier papers such as \citep{Calders_first_latent_attempt}, where the construction of a model containing latent variables is put forward. 

\newblock

Regularisation based techniques work by incorporating a measure of the algorithm's unfairness as part of the loss function \citep{Learning_Fair_Reps, fair_logreg_regularized, tick_cross_paper}. Parameters $\mathbf{w}$ are found by minimising an expression of the form:
\begin{equation}
    \mathbf{w} = \text{argmin} (L(\mathbf{w},X,Y) + D(\mathbf{w},X))
\end{equation}
where $L(\mathbf{w},X,Y)$ is a measure of classification inaccuracy, and $D(\mathbf{w},X)$ is a measure of the discrimination of the algorithm (typically a measure of how far it deviates from demographic parity) \citep{tick_cross_paper, Learning_Fair_Reps}. Thus, when the parameters of the model are adjusted, the contribution of those attributes which exhibit strong dependency on the protected will be reduced. These approaches have been very successful in minimising statistical measures of discrimination, and can achieve near demographic parity \citep{fair_logreg_regularized}. However, they require an arbitrary hyper-parameter setting that determines the weight given to the fairness metric in the loss function. Choosing this parameter inevitably involves a trade off between achieving fairness and maximising classification accuracy \citep{tick_cross_paper}. 

\newblock

Another problem with regularisation-based techniques is that they require an explicitly statistical treatment, as the training procedure must minimise an associated statistical error function; hence many of the criticisms which apply to statistical measures of fairness such as demographic parity also apply to the methods used to enforce them. Provided an algorithm satisfies a statistical criterion, the algorithm will be deemed fair regardless of how different individuals within the population are treated \citep{Fairness_through_awareness}. The authors of \citep{Fairness_through_awareness} introduce an additional fairness stipulation that similar individuals receive similar outcomes from the algorithm, thus preventing disparate treatment of similar individuals in order to satisfy a statistical goal. 

\newblock

Because of the way the error function is structured, regularised fair classifiers alter the parameters of the model to reduce the contribution of those attributes which are heavily statistically dependent on the protected attribute. However, if one or more of these attributes happens to be a good predictor of the final outcome, then the model may lose performance, and in general very low levels of disparate treatment are only achieved at a fairly significant cost to classification accuracy \citep{Learning_Fair_Reps}.
Moreover, it could be argued that certain individuals who score highly on one of the attributes (even though the remainder of their demographic generally doesn't) are unfairly penalised by a regularisation approach. As a concrete example, suppose members of population 0 generally score well on attribute $b$ but members of population 1 do not. A regularised classifier would reduce the contribution of variable $b$ to the final prediction $\hat{y}$ in an attempt to achieve demographic parity. Nonetheless, the few members of population 1 who score highly on attribute $b$ could argue they have been mistreated if they would have been rewarded a positive outcome by an `unfair' model but were instead awarded a negative outcome by a regularised model, despite the fact that in general their population benefits from the reduced weighting of attribute $b$. This is one argument in favour of `backing-out' the contribution of the protected attribute to the variable rather than ignoring the variable entirely. This issue does not appear to have been considered extensively in previous literature but it is worth noting and highlights the insufficiency of demographic parity as a fairness criterion. 

\newblock

These problems were partially addressed in \citep{Disparate_Impact_&_Historical}, where the authors instead attempted to regularize a classifier using a statistical measure called disparate mistreatment. This is a measure of the difference in the rates of incorrect negative outcomes between individuals in different groups. Such a technique is most appropriate when the label is what the authors refer to as a `ground-truth' criterion, which represents a known outcome, rather than a label allocated by humans that could be subject to historical biases \citep{Disparate_Impact_&_Historical, tick_cross_paper}.

\newblock

This paper focuses exclusively on achieving fairness through pre-processing, in a manner similar to that explored in \citep{Counterfactual_Fairness}, where we attempt to remove the contribution of the protected attribute to each variable prior to training time. Nonetheless, it is worthwhile to consider how  alternative techniques in the literature could be used instead. 

\newpage

\subsection{Tractable Inference and Sum Product Networks}

In this section, we briefly recap the fundamentals of SPNs. 
Sum Product Networks (SPNs) are a type of deep architecture  \citep{SPN}, and are essentially instances of  arithmetic circuits \cite{choi2017relaxing}. Mainly, they admit the tractable computation of inference queries in linear time \citep{SPN}. SPNs can be thought of as a network, where each leaf node is a tractable univariate distribution \cite{SPN_structure_learning}. A single query to an SPN involves setting the values for some or all of the leaf nodes, and the output can be interpreted as being proportional to the probability that the leaf nodes can take this set of values simultaneously; this amounts to computing an inference query of the form $P(X=x)$ where $X$ is some subset of all leaf-node variables. 

\newblock

SPNs can be defined recursively in terms of other sub-SPNs. Specifically, a new SPN can be created either from the product of several sub-SPNs with disjoint scopes or from a weighted sum of sub-SPNs with identical scopes, where the sum of the weights is unity \cite{SPN_structure_learning}. In the former case, the value of the root node in the SPN $S(\mathbf{x})$ is given by the product of the root nodes of the sub-SPNs:
\begin{equation} \label{eq: prod_node}
    S(\mathbf{x})=\prod_i S_i(\mathbf{x_i})
\end{equation}
where $x_i\cap x_j = \emptyset \hspace{1mm} \forall \hspace{1mm} i\neq j$. In the latter case, the value of the large SPN is a weighted sum of the root nodes of the sub-SPNs
\begin{equation} \label{eq: sum_node}
    S(\mathbf{x})=\sum_i w_i \times S_i(\mathbf{x}).
\end{equation}
Note that in the `sum-node' case, the scope of each contributing sub-SPN is identical to that of the composite SPN. 

\newblock

The value of the output $S(\mathbf{x})$ is proportional to the probability for the vector of attributes to take on that value, i.e. $P(x) \propto S(x)$. Crucially, SPNs have the property that it's possible to compute their partition function in linear time. This allows inference to be computed rapidly, since evaluating the partition function of a model is typically the limiting factor in terms of computational cost when making inference queries \citep{SPN}. The probability for a set of input attribute values $\mathbf{x}$ to occur is
\begin{equation}
    P(\mathbf{x})=\frac{S(\mathbf{x})}{Z(S)},
\end{equation}
where the partition function $Z(S)$ is given by
\begin{equation} {\label{eq: partition_function}}
    Z(S)=\sum_{\mathbf{x_i} \in \chi}S(\mathbf{x_i}),
\end{equation}
and $\chi$ is the set of all possible inputs to the SPN \citep{SPN}. This tractability property arises due to the restriction of the network to only product nodes or sum nodes built using  tractable univariate distributions, such as Gaussian random variables or Bernoulli random variables. By means a single pass of the network, conditional marginal can be computed, and more generally, the computational complexity of computing the partition function of the root node scales linearly with the number of edges in the SPN \citep{SPN_structure_learning}.

SPNs can be learned automatically from data; for example, \citep{SPN_structure_learning}  propose the ``LearnSPN'' algorithm schema.  In broad terms, LearnSPN attempts to identify approximately independent subsets of variables within a data set, and breaks these up into separate blocks of data. If there are no independent subsets, then the SPN instead looks to divide the data into blocks of similar instances, and then look for independencies within the resulting data set \citep{SPN_structure_learning}. A product node is created when the data is divided into independent subsets of variables and a sum node is created when the data is divided into separate blocks of similar instances. This procedure is repeated until the algorithm arrives at a leaf node, comprising a distribution over a single variable \citep{SPN_structure_learning}.

\newblock

To make the rationale behind this procedure explicit, consider the scenario where there is a product node at the origin. In this scenario, two or more sets of independent variables have been identified, and therefore one would expect the probability distribution to factorise into several corresponding distributions:
\begin{equation}
    P(X)=\prod_iP_i(X_i),
\end{equation}
where X is the set of all variables and each $X_i$ is a separate independent subset. It intuitively makes sense to insert a product node, since the resulting probability for some particular configuration of evidence $\mathbf{x}$ should be proportional to the product of the sub-SPNs. Recall that the output of the SPN is proportional to the probability for the evidence to take that value. 

\newblock

Meanwhile, subdividing the data into different instances can be regarded as being similar to conditioning on a particular attribute or set of attributes. Consider the root node again, only now assume that no subsets of independent variables can be found, and the data is instead subdivided into similar instances. The probability distribution over all instances $P(X)$ can be written as follows:
\begin{equation}
    P(X)=P(X_i|X_j)P(X_j)
\end{equation}
where $X=X_i \cup X_j$. The SPN is therefore divided up into different blocks of data on the basis of the values of the components of the set $X_j$ \citep{SPN_structure_learning}. The value of each distinct sub-SPN corresponds to the probability distribution over all variables $P(X_i|X_j)$ for some specific value set of subset of possible values of $X_j$, and the weights are given by the probability that the members of $X_j$ take that particular set of values - i.e. $w_j=P(X_j)$ \citep{SPN, SPN_structure_learning}.

\newblock

For the purposes of discussing the results of our experiments later in the report, it will be useful to briefly outline the mechanisms used by the LearnSPN algorithm to split the data into different subsets of variables and instances, and thereby construct the structure of the SPN. Independence was established using a G-test between two variables \citep{SPN_structure_learning}:
\begin{equation}
    G(x_1,x_2)=2\sum_{x_1}\sum_{x_2}c(x_1,x_2)\times \log(\frac{c(x_1,x_2)\times N}{c(x_1),c(x_2)}).
\end{equation}
Here, $N$ is the number of instances in the sample set or sub-sample set, and $c(x)$ is an counting function which returns how many times $x$ occurs, or, for $c(x,y)$ how many times that combination of $x$ and $y$ occurs \citep{SPN_structure_learning}. This pairwise test was done independently on every possible pairing of variables, and an independent subset of variables was identified if and only if all attributes within that subset were found to be approximately independent of all variables not contained in the subset. This is analogous to forming a graph where each vertex is a variable and edges are only allocated between variables if the pairwise G-test between two variables exceeds some threshold. Any isolated cluster may then be regarded as an independent subset of variables \citep{SPN_structure_learning}. It is worth noting at this juncture that the mutual information between training attributes and the protected attribute has been discussed in previous literature as a mechanism by which bias can be introduced \citep{Learning_Fair_Reps}. Therefore, the subdivision of attributes on the basis of their mutual information seems to be a sensible first step that remains in keeping with prior fair learning approaches.

\newblock

As mentioned previously, if this independence testing strategy proves unable to isolate two or more subsets of variables, then the data is instead divided into subsets of similar instances. The chosen method in \citep{SPN_structure_learning} was to cluster instances using hard incremental expectation maximisation, a technique derived from  \citep{Hard_Incremental_EM}. A naive Bayes model is assumed, or in other words it is presupposed that all variables are statistically independent given the cluster (in reality, this is naturally unlikely to be the case). The probability for a particular configuration of evidence $X$ is then modelled by:
\begin{equation}
    P(X)=\sum_i P(C_i) \prod_j P(X_j|C_i),
\end{equation}
where $i$ labels each cluster, and $j$ each variable. $P(C_i)$ is the prior probability for an instance to be assigned to cluster $i$ - these are used as the weights $w_i$ when constructing the SPN.

\newpage

\section{Experiment Methodology and Results}

\subsection{Theoretical Outline}

\subsubsection{The Rationale for Using SPNs}

The initial pretext for this experiment was to use SPNs to construct a counterfactually fair model. The motivation for the application of SPNs to problems in fair machine learning lies in their ability to efficiently compute conditional probabilities. Suppose we have a series of unprotected attributes $x_1,x_2,...,x_n$ and a series of protected ones $a_1,a_2,...,a_m$. Once an SPN is trained, we have the ability to calculate the probability of any set of values taken by the unprotected variables for some configuration of the protected attributes. We are thus able to compute inference queries of the form $P(x_1=\alpha,x_2=\beta,...,x_n=\omega|a_1=p,a_2=q,...a_m=z)$ in linear time using the SPN \citep{SPN}. As we have already discussed, many fair learning techniques are predicated on``backing-out'' the contribution of the protected attribute to the other variables prior to prediction \citep{Counterfactual_Fairness, Causal_Reasoning}. In principle, this technique should prevent the algorithm from becoming biased by exploiting the information about the protected attribute contained in the training variables. SPNs are therefore useful tools for learning conditional probabilities, allowing us to back out the contribution of the protected attribute to other variables.

\subsubsection{Identifying Dependent Criteria}

In \citep{Counterfactual_Fairness}, the authors demonstrate that any predictor trained solely on the non-descendants of a protected attribute in a Directed Graphical Model are guaranteed to satisfy counterfactual fairness with respect to the protected attribute. However, in the example presented in their paper, which concerns admission rates to law school for individuals of different ethnicities, the authors don't explicitly describe a method by which the non-descendants can be identified \citep{Counterfactual_Fairness}. Instead, they ``back-out'' the contribution made by the protected attribute to the training variables by assuming that the protected attribute makes a causal contribution to every training variable. Moreover, they assume the presence of a hidden variable, and assert that the hidden variable and the protected attribute each make an additive linear contribution to the training variable \citep{Counterfactual_Fairness}. As noted in the original paper, these are strong assumptions. One plausible improvement on this technique might involve establishing which attributes exhibit no statistical dependence on the protected attribute, negating the need to `invent' a hidden variable for each of these independent attribute. We could then safely train a model using only these attributes in the knowledge that it will not lead to biases, since none of them are statistically dependent on the protected attribute. 


\newblock

Earlier we suggested that SPNs can be used as a method of separating variables into two categories: those which are dependent on one or more protected attributes and those which are not. Recall that  the LearnSPN algorithm attempts to separate the data presented to it into subsets of variables, where all variables in one such subset may be statistically dependent other members of the same subset, but are independent of the variables in all other subsets \citep{SPN_structure_learning}. 

\newblock

In a directed graphical model, all non-descendents of a protected attribute must be statistically independent of it, provided that they are not ancestors of the protected attribute. Since protected attributes are typically fundamental personal attributes (e.g. sex, ethnicity), they cannot, in a `causal' sense, be a descendant of another attribute. To give an explicit example of what is meant by this, it's conceivable that salary could be influenced by gender but not vice-versa. Therefore, we take the view that any attributes which are statistically independent of a protected attribute are non-descendants of that protected attribute. 

\newblock

Consequently, if the SPN successfully subdivides the variables into several distinct subsets of variables $X_k$, where $k = 1,...,m$, then the distribution over all variables factorizes as:
\begin{equation}
    P(X)=\prod_{k=1}^{m}P_k(X_k),
\end{equation}
where
\begin{equation}
    X_k \subseteq X \text{ } \forall \text{ } k .
\end{equation}
If the protected attribute $a_p \in  X_j$, then $x_i \upvDash a_p \hspace{3mm} \forall x_i \not \in X_{j}$. Consequently, $x_i$ are non-descendants of $a_p$ and it is safe to train a fair classifier using these attributes. Clearly this approach can be extended to data sets with multiple protected attributes by only selecting variables belonging to subsets which contain no statistical attributes.

\begin{figure}[h!]
\centering
\begin{tikzpicture}[->,>=stealth',shorten >=1pt,auto,node distance=3cm,
                    thick,main node/.style={circle,draw,font=\sffamily\LARGE\bfseries}]

  \node[main node] (s1)  at (0, 0)  {$\times$};
  
  \node[main node, label=below:{\{$x_1,x_2,...,a_p$\}}] (r1) at  (-3, -2.5)   {$+$};  
  \node[main node, label=below:{$\{x_{l+1},...,x_n\}$}] (r2)  at (3, -2.5)  {$+$};
  \node[main node, label=below:{$\{x_k,...,x_l\}$}] (r3)  at (0, -2.5)  {$+$};
  
  \path[->]
   (s1) edge (r1)
   (s1) edge (r2)
   (s1) edge (r3);
\end{tikzpicture}
\caption{Illustration of variable separation by an SPN. The protected attribute $a_p$ is associated with the left-hand node only, so according to the statistical test here, the variables allocated to the other nodes are independent of the protected attribute.}
\label{fig:dirgraph}
\end{figure}
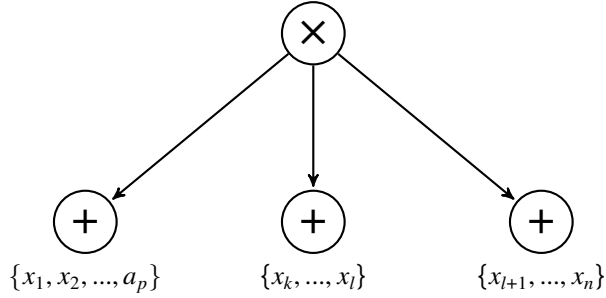

\subsubsection{Adjustment of Dependent Attributes}

Using SPNs, we have established a method which allows us to determine which variables are statistically independent of the protected attribute and are therefore `safe' variables to use when training a model. However, we must now consider how the remaining attributes can be modified to prevent the model learning biases. Since these variables are statistically dependent on the protected attribute, there is the potential for any model trained on them to learn biases by exploiting the information about the protected attribute contained therein. The strategy employed here is similar to that used in previous papers which attempt to establish fair treatment of individuals through pre-processing: namely, we attempt to remove the contribution which the protected attribute makes to each training variable \citep{Counterfactual_Fairness, Causal_Reasoning}. For continuous attributes, this presents no conceptual difficulty. If the contribution of the protected attribute to the variable is regarded as an additive constant, as it is in \citep{Counterfactual_Fairness}, then we can train a model ${f_i}(a_p)$ to predict the outcome of each dependent training variable $x_i$ given the protected attribute. The residual: 
\begin{equation} \label{eq:Adjust}
    \tilde{x_i}=x_i - f_i(a_p),
\end{equation}
may then be regarded as the training variable with the contribution of the protected attribute removed. Here, we use an SPN for the model $f$. This SPN should be trained using only the data from the subset of variables which exhibit dependency on the protected. Using an SPN in this way brings several advantages. Once the SPN is trained, we can run, in linear time, a multitude of inference queries of the form:
\begin{equation}
    P(x_i|a_p) \hspace{1mm} \forall a_p, (x_i \not{\ci} a_p)
\end{equation}
Moreover, if we have more than one protected attribute, but each protected attribute belongs to a separate subset of variables $X_k$, then the number of inference queries required grows linearly in the number of protected attributes. However, it should be noted that in the worst case scenario, where all protected attributes belong to the same subset, the number of queries required scales with $O(K^N)$, where $K$ is the number of values the protected attributes can take, and $N$ is the number of protected attributes. Provided the number of protected attributes isn't large this is unlikely to be a significant problem.

\newblock

When the protected attribute is not continuous, the appropriate procedure for adjusting the values becomes less obvious. For discrete attributes which still contain numerical values, it's still possible to train a model to predict the value of $\tilde{X_i}$ given the protected attribute or attributes, but now the resulting outcomes will likely no longer be discrete values, since the outcome of $f_i(a_p) = \sum_{x_i} x_i P(x_i|a_p)$ isn't constrained to be a discrete value. This is the approach that we elected to take here. 

\newblock

For qualitative categorical variables, the situation is even more problematic, as there is no associated ordering to the values the attribute can take. One solution, of course, is to simply remove from the data set any categorical variables which exhibit any statistical dependence on the protected attribute. However, this could severely compromise the accuracy of the final model. We attempt here an alternative method, whereby the probabilistic contribution of the protected attribute is removed using the following procedure. Assume a categorical attribute $x$ can take unordered values in the set $C = \{c_1,c_2,...,c_n\}$. Firstly, using the SPN, we compute the conditional distribution over all attributes given the protected and express it as a vector:
\begin{equation}
    \vec{p}(x|a_p=a) = [p(x=c_1|a_p = a), p(x=c_2|a_p = a), ... , p(x=c_n | a_p = a)] \hspace{1mm} \forall a.
\end{equation}
Subsequently, the protected attribute is one-hot encoded such that $c_i \rightarrow \vec{1}_i$, where $\vec{1}_i$ is a unit vector with 1 in the ith position and 0s everywhere else. Then the adjusted value of x is given by:
\begin{equation}
    \tilde{x}(x=c_i, a_p=a) = \vec{1}_i - \vec{p}(x|a_p = a)
\end{equation}

\newblock

The solution proposed here for dealing with qualitative categorical attributes is not without its shortcomings. If the machine learning model is given sufficient flexibility, it may be able to reacquire explicit information about the value of the protected attribute, since the new position of the adjusted value $\tilde{x}$ in the n-dimensional space of all possible encodings is explicitly dependent on the protected attribute.

\newblock

If the presence of a particular categorical attribute is a good indicator of the outcome of the training labels, and the model is constrained to learn a distribution which increases or decreases monotonically with the new adjusted values $\tilde{x}$ (for example a logisitic regression classifier), then the procedure presented here above should improve make the model fairer. The values presented to the model are adjusted in such a way that the demographic more likely to present that attribute is awarded a lower value, and therefore that demographic should receive an associated penalty, thus making the classifier fairer. However, if the same data is presented to a more complicated machine learning such as a neural network, which is capable of identifying non-linear patterns in the data, then the correctional procedure may actually work to compromise fairness. Nonetheless, we investigate what effect this correctional procedure has on the outcome of the model whilst bearing this caveat in mind.

\subsection{Initial Experiments with One Protected Attribute}

\subsubsection{Data Set 1: German Credit}

Our initial experiments were carried out on a data set compiled by an academic at Hamburg university and made public in 1994 \footnote{Full details and data are available from the UCI machine learning repository: https://archive.ics.uci.edu/ml/datasets/statlog+(german+credit+data)}. It contains information about the credit-worthiness of individuals (the target classification criterion) as well as a host of financial and personal information. In total the data set comprises 1000 individuals, with 20 different training attributes and one binary training label associated with each individual \citep{German_Credit_Source}. All the individuals in the data set, were classed as either having `bad' or `good' credit risk. The task is to predict the outcome of this credit rating from the training data.

\newblock

There were several reasons for selecting this data set as a testing ground for novel techniques. Firstly, there are several training variables which could conceivably be classed as attributes worthy of protection, including an individual's age and whether or not they were a foreign worker. However, for the sake of simplicity in our initial experiments, we elected to use sex as the protected attribute, partly because it's a simple binary criterion, but also because both `male' and `female' are present in large proportions. By contrast, only a very small fraction of individuals were foreign workers. Initially, we are therefore only interested in sex-based discrimination, and we aim to predict the credit score in each case.\footnote{It is interesting to note that \citep{German_Credit_Source} refers to the credit-worthiness target criterion as a `credit-risk'. Since it doesn't represent a `ground-truth' criterion (e.g. defaulted on loan) but rather an assessment of an individual's risk, there is the possibility that the label themselves may have been subject to human biases when they were allocated, and we choose to treat them as such.}


\newblock

Furthermore, the data contains a mixture of categorical, discrete numeric and continuous numeric attributes, and is therefore fairly typical of the types of data one would expect to see in the real world. Ideas such as counterfactual fairness more readily lend themselves to data sets comprising purely continuous numerical data, and indeed the example in \citep{Counterfactual_Fairness} comprised only numerical attributes. Overall, of the initial 20 training attributes, 13 were qualitative categoricals (including the protected attribute), a further 4 can be considered discrete numerical variables, and the final 3 are most readily treated as continuous.


\newblock


\newblock

\begin{figure}[h!]
\centering
\includegraphics[scale=0.5]{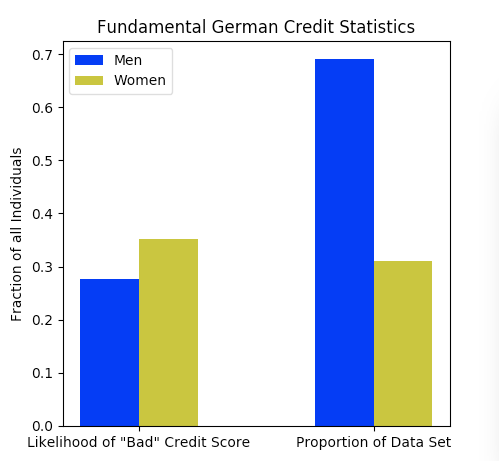}
\caption{Approximately two thirds of the population is male, and women are 27\% more likely than men to be awarded a ``bad'' credit score.}
\label{fig:German_Credit}
\end{figure}


\subsubsection{Data Pre-Processing}

The entire data set of 1000 instances was first randomly divided into a test set (containing the data from 200 individuals) and a training/validation set comprising the remaining 800 individuals. This was subsequently sub-divided into 5 blocks of 160 instances for a k-fold cross validation procedure. Each of the 5 training sets, comprising 640 applicants, served as a separate training set for the SPN. The binary sex attribute which we treat as `protected' was originally further subdivided into additional categories on the basis of marital status. This additional feature was ignored, and instances were just classified as being either male or female. 

Qualitative and discrete numerical attributes were all one-hot encoded, such that a single attribute was now treated as a list of several separate attributes. Meanwhile, continuous values were each allocated to one of 10 equally spaced bins, and then subsequently one-hot encoded. This procedure meant that the SPN is fed a total of 93 different attributes, since each original attribute (assuming it wasn't binary to begin with) is now encoded by several different variables. (SPN extensions for hybrid domains, such as \cite{molina2018mixed,bueff2018tractable}, might offer more elegant alternatives.)

%

\subsubsection{Attribute Separation and Adjusting for Dependency}

Initially, we used a G-test threshold of 0.0001, which was one of the values used when performing SPN structure learning in \citep{SPN_structure_learning}. The purpose of this first run was not to train an SPN per-se but instead to ascertain whether it was possible to identify a subset of variables which exhibited strong statistical dependence on the protected attribute. The original structure of the SPN was modified so that the first task was to attempt to identify mutually independent subsets rather than to subdivide the data into blocks of different instances. If the algorithm was able to successfully find mutually independent subsets of variables, then these subsets were written to a  file, from which the subset containing the protected attribute could be identified and isolated from the remaining variables. At the first stage of separation, the SPN successfully identified that all attributes arising from a single one-hot encoding should be allocated to the subset of variables.

%

This procedure was carried-out for each of the 5 different training sets available (i.e. 5 overlapping subsets of the 800 individuals, each comprising 640 instances). Three of the experiments resulted in identical independent subsets of variables being identified, and a fourth isolated the same subset but also included one additional discrete numeric attribute. The independent set with the protected attribute arising from the fifth and final training set also contained all the variables contained in the first four `dependent' subsets, and also included several other variables besides, meaning that a smaller subset of variables were regarded as being independent of the protected attribute.

\newblock

Subsequently, a second SPN was trained on all the variables belonging to the subset containing the protected attribute. This allows us to construct a model predicting the outcome of a variable given the protected attribute, and thus construct the models $f_i(a_p)$ representing the contribution of the protected attribute to the training variables in equation \ref{eq:Adjust}. 
\subsubsection{Training Models}

Only 640 instances are available for training in each case, each of which comprises 56 attributes (this number includes every value arising from the one-hot encodings for categorical attributes). In view of the size of the data set, we elected to train three fairly simple models: a Logistic Regression classifier, a Gaussian Naive Bayes model, and a Support Vector Machine. All of these models are relatively computationally inexpensive to train, and have a limited number of parameters. This is likely to prevent the overfitting one might expect to encounter if training a more complicated architecture such as a deep neural network \citep{ML_Book}. Also, these models have been used on previous occasions when conducting experiments on this data set \citep{Learning_Fair_Reps, Disparate_Impact_&_Historical}, and therefore remain useful for comparability purposes. Finally, we should stress that the purpose of our experiment is not necessarily to find the best performing classifier, but rather to ascertain how our pre-processing procedure affects classification accuracy and fairness across a range of different classifiers. Therefore, we can assess whether the techniques for training variable pre-processing are robust across a range of different models. All models were implemented using Scikit-learn. 

To assess the difference in classification accuracy made by our pre-processing technique, every model was trained on 4 different versions of the data set. The first was the ``unfair'' data set which contained just the raw data and the protected attribute. The second model also contained the raw data, but now with the protected attribute removed; this technique is an implementation of the ``fairness through unawareness'' approach discussed in section 3.1.1. This may be regarded as an appropriate fairness baseline. Next, we trained two different fair models. The first of these was trained on the fair data, which had been adjusted using the technique described above, and validated/tested on data adjusted in the same way. In the second fair technique, the model was allowed to see a modified version of the ``unfair'' data at training time, but was subsequently validated/tested on data where the conditional dependency on the protected attribute was removed.

\paragraph{Fair Learning Technique 1}
Consider a predictive model $g$ with trainable parameters $\mathbf{w}$, such that the prediction of any given point is given by
\begin{equation}
    \hat{y} = g(\mathbf{x}, \mathbf{w}).
\end{equation}
In the first instance, these parameters are set by minimising a Loss function $L$, which is a function of the adjusted data, $\tilde{X}$ and training labels $Y$.
\begin{equation}
    \mathbf{w}_{1} = \text{argmin } L(\tilde{X}_{\text{train}}, Y, \mathbf{w}_1)
\end{equation}
$\tilde{X}$ has been adjusted according to the technique described in equation \ref{eq:Adjust}. All predictions carried out on the validation set are then achieved on data which has been adjusted via the same procedure.

\paragraph{Fair Learning Technique 2}
This method demands that the model is first trained on data where the effects of the protected attribute have  not yet been compensated for. The new ``centred'' data, referred to as $X^{C}_{\text{train}}$, is calculated by subtracting a weighted sum of the contribution of all protected attributes $f_i({a_p})$.
\begin{equation}
    \tilde{\mathbf{x}}^{C}_i = \mathbf{x}_i - \sum_{a_p}P(a_p) f_i(a_p).
\end{equation}
Subsequently, the parameters $\mathbf{w}_2$ are set to minimise the same loss function used in technique 1, only now the ``centred'' data is used in lieu of the ``adjusted'' data. Validation tests are then carried out on the ``adjusted'' data, where the contribution of the protected attribute has been removed, much as they were in the previous technique.

\newblock

By testing these two separate fair methods, we aim to ascertain whether the optimal strategy is to remove the contribution of $a_p$ prior to training, or to train the model on unfair data but remove the contribution of $a_p$ before testing. There is a risk that technique 2 could compromise classification accuracy, since the model is now predicting data adjusted in a different way to that which it was trained on. On the other hand, technique 2 mitigates our earlier concerns about the potential for the model to reacquire explicit information about the protected attribute because of the way we adjusted the qualitative attributes.

\newblock

In all cases, each training attribute was scaled to lie in the interval $[0,1]$ prior to training in order that any regularisation term would have a similar effect on the magnitude of all weights.

\paragraph{Logistic Regression}
The Logistic Regression classifier assigns a probability that each test point is in class 1:
\begin{equation}
    g(\mathbf{x}) = P(y=1|\mathbf{x}) = \frac{1}{1+\exp(-(\mathbf{w}^T \mathbf{x}+b))}
\end{equation}
The weight parameters $\mathbf{w}$ are trained using a gradient descent procedure to minimise the loss function, which consists of the Negative Log Likelihood plus a regularisation term $\lambda \mathbf{w}^T\mathbf{w}$ that penalises large weight values for $\mathbf{w}$. The regularisation term $\lambda$ by running trials for multiple different values of $\lambda$ on the validation set. Every one of the 4 data sets was tested for 4 different values of $\lambda$ ($\lambda = \{1, 0.5, 0.2, 0.1\}$. In general, it was found that smaller values of $\lambda$ slightly increased classification accuracy, but this also came at a cost of a slight increase in disparate treatment in all four models. As a compromise, $\lambda = 0.5$ was selected for the final model.
    
\paragraph{Gaussian Naive Bayes}

A Gaussian Naive Bayes model assumes that the training variables of instances belonging to a particular class $c$ are mutually independent and Gaussian distributed, such that the probability density over a single feature $x_j$ is given by:
\begin{equation}
p(x_j|y = c) = \prod_{j} N (x_j|\mu_j^{c},\sigma_j^{c}).
\end{equation}

Using Bayes rule, $P(y|\mathbf{x}) = p(\mathbf{x}|y)P(y)/p(\mathbf{x})$, the classifier can then calculate the probability that the instance with feature vector $\mathbf{x}$ belongs to a particular class. Note that the probability density $p(\mathbf{x}) = \sum_c(p(\mathbf{x}|y=c)P(y=c)$, where $P(y=c)$ represents a prior for an instance to be in class $c$ before any information about the feature vector $\mathbf{x}$ is known. The parameters $\mu_i$ and $\sigma_i$ are set from the training data. This assumes all training variables are mutually independent of one another, which we already know not to be true. However, for large data sets, independence assumptions reduce computational cost, since there is no need to calculate correlations between all variables if independence is assumed.

\paragraph{Support Vector Machine}

The final model trained here was a Support Vector Machine with a soft margin and a radial basis function kernel. The outcome of a new test point is given by:
\begin{equation}
    y(\mathbf{x}) = \text{sign}\left(\sum_{i=1}^{N}\alpha_iy_i\exp(-\gamma (\mathbf{x_n}-\mathbf{x})^{T}(\mathbf{x_n}-\mathbf{x}))+b\right).
\end{equation}
The parameters $\alpha_i$ are found by maximising the expression:
\begin{equation}
    \sum_{i=1}^{N} \alpha_i - \frac{1}{2}\sum_{i,j = 1}^{N}\alpha_i\alpha_jy_iy_j\exp(-\gamma (\mathbf{x_n}-\mathbf{x})^{T}(\mathbf{x_n}-\mathbf{x}))
\end{equation}
subject to the constraints,
\begin{equation}
    \sum_{i=1}^{N}\alpha_iy_i = 0 \hspace{5mm} \text{  and  } \hspace{5mm} 0 \leq \alpha_i \leq C \text{ } \forall i
\end{equation}
The formulation of these expressions is not discussed here, but further details can be found in \citep{ML_Book}. The important point is that, unlike for a logistic regression classifier, the decision boundary is no longer a linear function of $x$, but will now in general be a manifold embedded in the space of all features. The kernel coefficient $y$ and upper-bound parameter $C$ were set by comparing classification accuracies on held-out validation sets using a 5-fold cross validation scheme. The optimal value pairing was found to be $\gamma = 0.1$ and $C = 1$. Scikit-learn's implementation also comes with a probabilistic estimator for test points, which allocates a probability as well as a class to each individual tested. We use this feature when considering a probabilistic measure of fairness.

\subsubsection{Model Assessment and Analysis}

\begin{figure}[h!]
\centering
\includegraphics[scale=0.5]{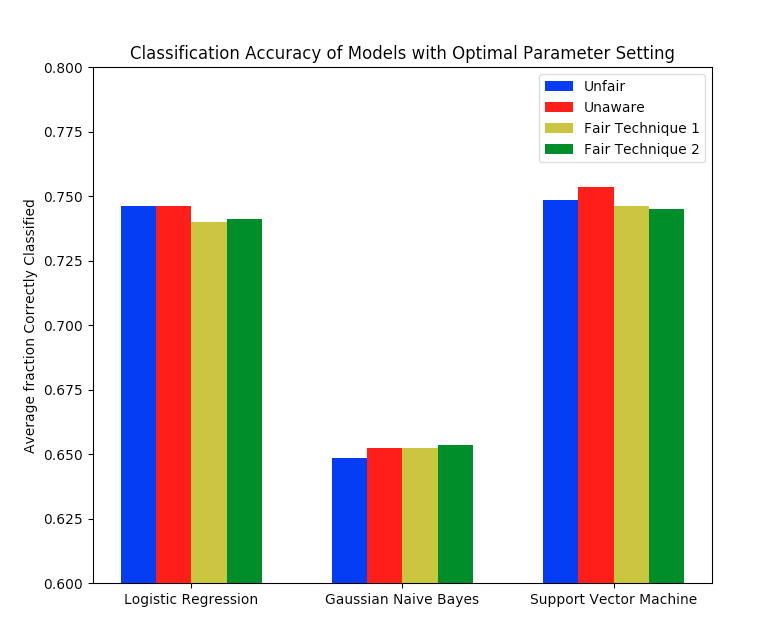}
\caption{Average Comparative Classification Accuracy for 3 different models (Note y-axis scale). }
\label{fig:German_Accuracy}
\end{figure}

Every model was assessed on the basis of a performance criterion (percentage of correctly classified individuals), and two fairness criteria. We consider firstly the discrepancy in classification accuracy across models (shown in figure \ref{fig:German_Accuracy}). The classification accuracy of all 4 models on the held out validation sets was broadly comparable. For both the logistic regression and the support vector classifiers, the ``unfair'' and ``unaware'' models generally performed marginally better than the adjusted models.

As discussed in the background section, adjusting to compensate for fairness frequently compromises the efficacy of the classifier \citep{Learning_Fair_Reps, Equality_of_opportunity}, so this was not unexpected and the fact that this effect is only slight is very encouraging. For example, the average classification accuracy for the logistic regression classifier on the ``fairness through unawareness'' data is \textbf{0.746}, whilst for the ``fair'' data, the classification accuracy is reduced slightly to \textbf{0.740}. There is no statistically significant difference between the performance of fair technique 1 and fair technique 2.

Every single Gaussian Naive Bayes model performed far worse than either of the other classifiers. One plausible explanation for this is the presence of strong correlations between different variables leading to poor classification accuracy under the naive Bayes assumption of independence. It's apparent that the Support Vector Machine marginally outperforms the Logisitic Regression classifier, most likely because the logisitic regression classifier must parition the data along a flat hyperplane, whereas the Support Vector Machine with a Gaussian kernel has the flexibility to learn a non-linear partition.

\newblock

\newblock

The similarity in performances for all four models is very encouraging: the best performing fair model under-performed the best performing unaware model by an average of only 0.75\%. Previous approaches using `fair regularization' have led to significant reductions in classification accuracy of over 3\% on the same data set \citep{Learning_Fair_Reps} (although it should be noted that the protected criterion was age and not gender in this instance, so the results are not directly comparable). Also, fair pre-processing strategies have previously encountered criticism because of concerns about the effect of data modifications on classification accuracy \citep{tick_cross_paper}, so it is interesting to see that in this instance the number of false classifications is only marginally reduced.

\newblock

We assess the fairness of the model on the basis of two criteria. The first of these is the difference in bad classification probability between the two genders, and we refer to this as disparate classification impact.

\begin{figure}[h!]
\centering
\includegraphics[scale=0.38]{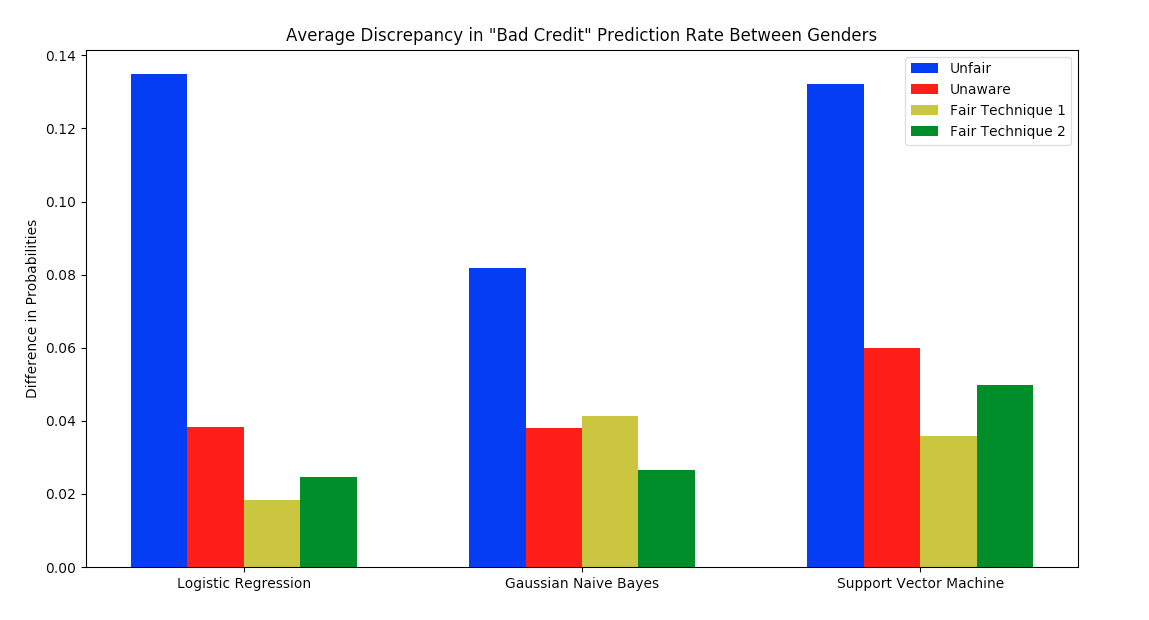}
\caption{Disparate Classification Impact for 3 classifiers and 4 different approaches to fair treatment.}
\label{fig:German_Fairness}
\end{figure}

\begin{definition}{Disparate Classification Impact}
    \newline
    For a classifier which makes predictions $\hat{y} = g(\mathbf{x})$, where $\mathbf{x}$ is the associated feature vector, the disparate classification impact is given by:
    \begin{equation}
        D_{\text{class}}(\hat{Y}) = \sum_{i=1}^{N} \left(\frac{I(\hat{y}^i = 1)I(a_p^{i}=\text{Female})}{N_{\text{Female}}} - \frac{I(\hat{y}^i = 1)I(a_p^{i}=\text{Male})}{N_{\text{Male}}} \right)
    \end{equation}
    where $\hat{Y}$ is the vector or all predictions made on the validation/test set, $N_{\text{Female/Male}}$ is the number of female/male members in the sample and $I(b)$ is an indicator function for a boolean:
    \begin{equation}
        I(b) = \begin{cases}
    1, & \text{if $b$ is True}.\\
    0, & \text{otherwise}.
  \end{cases}
    \end{equation}
    This is ultimately just a measure of the difference between the instance rates of `bad credit' predictions for women and men - i.e. $P(\hat{y}=\text{bad}|\text{female}) - P(\hat{y}=\text{bad}|\text{male}) $. This is equivalent to the discrimination criterion used as a measure of fairness in \citep{Learning_Fair_Reps}. A comparison of disparate classification treatment for the models trained here is shown in \ref{fig:German_Fairness}.
\end{definition}

The second criterion is defined in terms of a sum over the individual probability that a specific member of a demographic group is bad, and can therefore be viewed as a Bayesian analogue of the first criterion.

\begin{figure}[h!]
\centering
\includegraphics[scale=0.38]{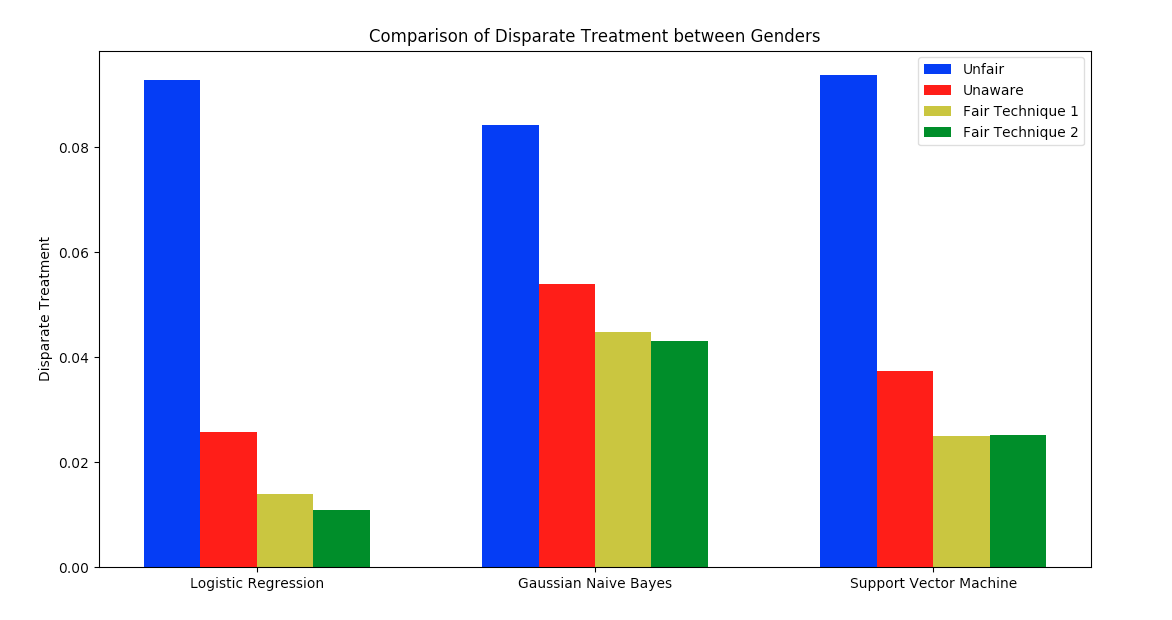}
\caption{Disparate Probabilistic Impact for 3 different classifiers and 4 different approaches to fair treatment.}
\label{fig:German_Disparity}
\end{figure}

\begin{definition}{Probabilistic Disparate Impact}
    \newline
    For a classifier which outputs a probability for any test point to be in class 1, such that  $\tilde{g}(\mathbf{x})=P(y=1|\mathbf{x})$, the disparate impact $D(X)$ is taken to be the difference in the normalized sum over all probabilities for males and females. It can also be described as the difference in the expectations of the probability that the outcome is positive outcome given the protected attribute. $X$ is the matrix containing all $N$ attributes. Formally, this may be written as follows:

    \begin{multline}
    D(X) = \sum_{i=1}^{N}g(\mathbf{x_i})\hspace{1mm}\left(\frac{I(a_{p}^{i} = \text{Male})}{N_{\text{Male}}} - \frac{I(a_{p}^{i} = \text{Female})}{N_{\text{Female}}}\right) \\ = \mathbb{E}(P(y=1|a_p=\text{female})) - \mathbb{E}(P(y=1|a_p=\text{male}))    
    \end{multline}

This is a broader definition of disparate impact. For example, if the classifier expressed a high degree of confidence in giving positive outcomes to males, but a low degree of confidence when allocating to females, then this isn't truly fair. Such a discrepancy will be reflected in this measure but not in the first one. Note this metric was used as part of the regularisation function in \citep{Learning_Fair_Reps}. The graph showing a comparison of disparate probabilistic impact is shown in figure \ref{fig:German_Disparity}.
\end{definition}

Since our data contains qualitative attributes, we cannot compute a set of counterfactual values as done in \citep{Counterfactual_Fairness}, because we cannot consider whether a qualitative attribute would have been different in the case that the individual in question were female rather than male or vice versa. This sort of approach is inherently limited to continuous attributes, where some assumption about the quantitative contribution of the protected attribute can be made. Nonetheless, we use the above discrimination metrics to measure deviation from demographic disparity, since this is regarded as the most appropriate metric when the labels are manually allocated \citep{Disparate_Impact_&_Historical, tick_cross_paper}.

\newblock

In every case, our data manipulation techniques both led to a consistent reduction in disparate probabilistic impact relative to the fairness through awareness model. This trend was consistent across all models regardless of hyperparameter settings (not shown here due to minor variations of the existing plots).  Consequently, we can conclude that these methods represent an improvement on fairness through unawareness. The average disparate classification impact was also reduced in most cases. (However, for the Gaussian Naive Bayes model, Fairness Technique 1  leads to a slight worsening in disparate classification impact, despite an improvement in disparate probability impact.) This illustrates that the modification of data prior to either training or testing can make a model fairer.

\newblock

For the best performing models (logistic regression and support vector machine), fair technique 1 tended to present with lower levels of disparate classification impact than fair technique 2, and therefore we select fair technique 1 as our final model to use on the test set. This potentially implies that it is better to remove the contribution of protected attributes prior to training the model. If the model is first trained on the raw data and then the validation data is adjusted, the classifier may have already learned to discriminate on the basis of variables which correlate strongly with the protected attribute \citep{Calders_first_latent_attempt}. The additive model assumed in equation \ref{eq:Adjust} may be insufficient to fully remove the contribution of the protected attribute to the testing data, and this may explain why technique 2 illustrates marginally higher disparate classification in this case.

\newblock

It is also apparent that the support vector classifier is ``less fair'' than the logistic regression classifier, particularly when comparing scores for the fair data sets. The average disparate classification when the logistic regression classifier was trained on the fair technique 1 on data was $\mathbf{0.018}$. When a support vector machine was trained on the same data, the result was $\mathbf{0.036}$, which is only a little lower than the disparate impact for the logistic regrssion model trained on the ``fairness through unawareness'' data. This is most likely attributable to the inherently linear data partition imposed by the logistic regression model. We have attempted to remove the contribution of the protected attribute to each variable by constructing an additive model (equation \ref{eq:Adjust}) as suggested in \citep{Causal_Reasoning} for linear models. However, models which are capable of learning to discriminate on nonlinear combinations of features may not have the full contribution of the protected attribute removed, since $\mathbb{P}(x_1 x_2|a_p) = \mathbb{P}(x_1|a_p)\mathbb{P}(x_2|a_p) $ is only guaranteed to hold if $x_1 \ci x_2 | a_1$. For $x_1 \not{\ci} x_2 | a_1, (x_1 - \mathbb{E}(x_1|a_p))(x_2 - \mathbb{E}(x_2|a_p)) \neq x_1 x_2 - \mathbb{E}(x_1 x_2|a_p)$ in general, and so the classifier may be left with some residual information about the protected attribute if it is capable of learning on variable pairings. The SVM partition boundary is not constrained to be linear and therefore this may plausibly be the cause of the additional discrimination manifested in the more complicated classifier. In view of the increase in disparate treatment associated with the SVM classifier, we elected to use the regularized logistic regression classifier as our final model to test on the test set. The final classification accuracy was $\mathbf{0.73}$, with a disparate classification impact of only $\mathbf{0.019}$.

\newblock

It should be noted at this juncture that regularisation-style intra-processing techniques have managed to achieve lower levels of disparate impact than we achieved with this technique \citep{Learning_Fair_Reps}. However, in their `fairest' model, which almost achieved demographic parity while using age as a protected criterion, the impact on classification accuracy was almost 8\%. Again, we stress that age is not regarded as a protected attribute in this instance and therefore the results aren't directly comparable. Nonetheless, further work could look to compare the technique considered here with established regularisation techniques and thereby assess whether the same reduction in disparate mistreatment achieved in this model has a better or worse effect on classification in a regularised model.

\newpage

\section{Percentile Equivalence}

\subsection{Percentile Equivalence with Continuous Attributes}

\subsubsection{Issues With Previous Approaches}

Thus far, we have attempted to remove the contribution of protected attributes to training variables by assuming they make a constant contribution. In our initial experiments on the German Credit data set, we trained a model $f_i(a_p)$ to predict the value of variables $x_i$ from the protected attribute, and then took this to be the contribution of the protected attribute to the training variable. Previous papers with a pre-processing oriented approach to machine learning have followed similar procedures \citep{Counterfactual_Fairness, Causal_Reasoning}. However, such adjustments are only fair and effective if the contribution of the protected attribute to the other training variables is additive and linear \citep{Causal_Reasoning}.

\newblock

Whilst these approaches work well for those individuals who lie close to the mean of their particular demographic group (i.e. ``typical'' individuals), the example below demonstrates how individuals at the extremities can be unfairly penalised by such an approach. This is particular true if the shape of the conditional probability distribution varies significantly between different demographic groups.

\subsubsection{Artificial Example for Continuous Data}


Consider the following hypothetical setup, which may be regarded as a simplified version of the German Credit data discussed earlier. Suppose a company is attempting to allocate loans to a number of individuals, and that they wish to do this in a fair way. They have a data set of $N$ prior individuals, which are fully characterised by 3 pieces of information: their salary $x$, a binary protected attribute $a_p$, and a credit rating label allocated to them denoted $y$. The salary $x$ is considered continuous. The protected attribute $a_p$ denotes which of two particular demographic groups an individual within that population belongs to. It is assumed that all individuals belong to one of two different groups, labelled V and W (these variables could, for example, encode different sexes or ethnicities). The aim is to train a fair predictor $\hat{y}$  which gives the probability that a particular individual will default on their debt without discriminating based on $a_p$ - that is, individuals in both V and W should be treated fairly.

\newblock

Salary $x$ is considered to be a good predictor of credit score $y$. However, the probability distribution over salary is different for the two demographic groups W and V:

\begin{equation}
    P(x|a=W) \neq P(x|a=V).
\end{equation}
In general, there is no reason to assume that the standard deviation, skew or kurtosis of these two distributions are necessarily similar. 

\newblock

\begin{figure}[h!]
\centering
\includegraphics[scale=0.6]{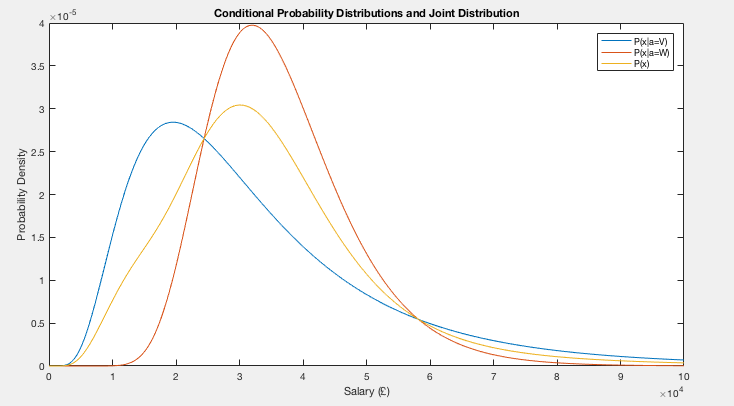}
\caption{Comparison of Distributions over Salary for groups W and V, and the joint distribution over all individuals.}
\label{fig:conditional_dist}
\end{figure}

In this example, we assume that the salaries for both groups are log-normally distributed (and therefore fully characterized by their standard deviation and mean), but that the W’s have a higher average salary and lower standard deviation. Therefore, W’s are likely to have a higher salary in general, but are much less likely to earn very high salaries. This represents a complex unfair world, where members of group W are likely to receive better remuneration in general, but members of group V still control positions in the highest paying industries. Whilst this is not a precise analogy, the situation shown here is not overly dissimilar to UK gender-pay disparities. However, in that scenario, men both earn more on average and have a higher standard deviation, so the discrepancy between the sexes is exacerbated for the highest earning individuals \citep{Salary_bot}. A graph showing the two distributions and the joint distribution is shown in figure \ref{fig:conditional_dist}.



\newblock

\begin{figure}[h!]
\centering
\includegraphics[scale=0.5]{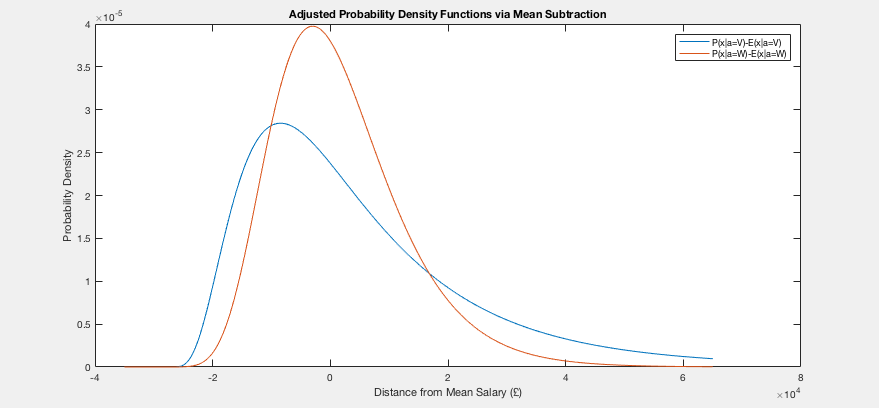}
\caption{Subtracting the mean of each distribution leads to greater similarity close to the mean, but exacerbates the discrepancy at high salaries.}
\label{fig:adjusted_dist}
\end{figure}

In this hypothetical example, simply subtracting the mean of the salaries results in unfair punishment for the highest earning members of group W. Since the mean salary of W is higher, they will receive a stiffer adjustment penalty, despite the fact that they are much less likely to earn very high salaries than members of V. Figure \ref{fig:adjusted_dist} shows the adjusted distributions, where the means have been subtracted: despite the distributions matching more closely around the mean, the situation in the high-salary region is now even more biased in favour of members of group V than it was before. 

\newblock

Moreover, the lower paid members of group V are also unfairly punished. The mean salary of group V is lower but still comparable to that of individuals in group W. Since members of V are much more likely to earn a very low salary than their counterparts in group W, the  adjustment is insufficient to compensate for their low earnings.

\newblock

Whether the model is unfair towards group W or group V is dependent on the risk attitude of the loan-issuer. Let’s suppose that the loan issuer is highly risk averse and only gives loans to the top 10\% of applicants (i.e. those with the highest adjusted salaries). This means that members of group V will already be over-represented, and subtracting the mean will make the situation even worse. This can be seen in figure  \ref{fig:adjusted_dist}. Thus we have demonstrated that in certain scenarios, fixed linear adjustments can unfairly penalize certain individuals.

\subsubsection{Resolving the Issue: Fairness through Percentile Equivalence}

In scenarios such as these where the unfairness inherent in the world is more complicated than a simple preference for one group over the other, we require a more nuanced technique to eliminate bias than simply subtracting off a constant contribution made by the protected attribute to the training variables. The solution presented here can be thought of as ``fairness through percentile equivalence''. The central premise of this adjustment is that individuals at the same percentile of their respective distributions should be treated identically. Therefore, the training variables are adjusted in such a way that individuals with attributes at the same percentile of their own distribution are allocated adjusted variables with the same value. This modification results in a ``fair world'' where the adjusted variables $\tilde{x}$ are independent of the protected attribute $a_p$.

\newblock

This is achieved as follows: the percentile $\beta$ of any individual in group $a_p=W$ can be calculated as follows:

\begin{equation}
    \beta_W=\int_{0}^{X_{true,V}}P(x|a_p=W)dx
\end{equation}
An analogous expression holds for $\beta_V$. By the definition outlined earlier, if $\beta_V=\beta_W$, then the adjusted variables $X_{true,V}=X_{true,W}$ should be identical. This can be achieved by defining a ``fair'' distribution $P_{fair}(x)=P(x)$, which is simply the joint distribution over all possible values of the protected attribute $a_p$:
\begin{equation}
    P_{fair}(x)=\sum_{a_p}P(x|a_p)P(a_p).
\end{equation}
Subsequently, we define the adjusted ``fair'' training variable $X_{adjusted}$ such that:

\begin{equation}
    \int_{0}^{X_{true~W}}P(x|a_p=W)dx=\beta_W=\int_{0y}^{X_{true~V}}P(x|a_p=V)dx=\beta_V=\int_{0}^{X_{adjusted}}P_{fair}(x)dx.
\end{equation}
It is immediately apparent from this that individuals at the same percentile of their respective conditional distributions will be treated identically.

\newblock

Returning to our artificial example, we can now guarantee that the model will be indifferent with respect to both $W$ and $V$ regardless of the model used to select applicants, provided it is trained using the new $X_{adjusted}$ variables, and that these are used when making predictions. 

\subsection{Application to a Simple Data Set}
We elected to test the percentile equivalence technique on a simple data set to gauge whether it mitigated the unfair treatment of individuals at the extremities of their respective distributions in the way we expect. The data set in question contains information about the performance and ethnicity of 21,790 students admitted to law school in the United States \citep{lsac_information}. The GPA and LSAT admission scores are provided as training attributes, and the task is to predict the first year average (FYA) of each student without discriminating on the basis of the protected attribute. The protected attribute is a binary indicator of ethnicity, where 0 denotes a White or East-Asian applicant, and 1 denotes a Hispanic or Black applicant \citep{lsac_information}. This data set was also used in \citep{Counterfactual_Fairness}. To establish whether the percentile equivalence idea makes a discernible difference to applicants at the tail end of the distributions, we compare the pre-processing method described above with the ``constant correction'' model used in \citep{Counterfactual_Fairness}. 

\newblock

\begin{figure}[h!]
\centering
\includegraphics[scale=0.5]{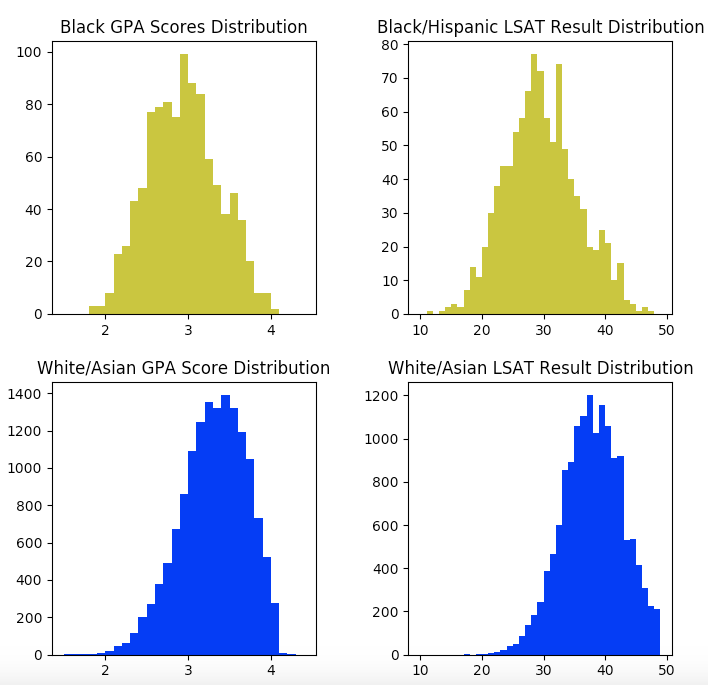}
\caption{Comparison of LSAT and GPA scores for White/Asian applicants and Black/Hispanic applicants.}
\label{fig:Histograms}
\end{figure}

Black/Hispanic applicants were found to have performed worse in admissions test on average than their White/Asian counterparts, as shown in figure \ref{fig:Histograms}. More significantly from the perspective of the technique described above, the scores of Black/Hispanic applicants were also found to have a higher standard deviation, and therefore there is a risk that the lowest scoring Black/Hispanic applicants will be treated unfairly by any pre-processing technique which assumes a constant contribution from the protected attribute. These discrepancies should be compensated for prior to training in order for the model to be fair. The statistics for different applicants are shown in the table overleaf.

\begin{table}[h!]

\begin{tabular}{ |p{3cm}|p{2cm}|p{3cm}|p{2cm}|p{3cm}|  }
 \hline
 \multicolumn{5}{|c|}{Comparison of fundamental statistics} \\
 \hline
 Demographic Group & GPA Mean & GPA Standard Deviation & LSAT Mean & LSAT Standard Deviation \\
 \hline
 Black/Hispanic Applicants & 2.890 & 0.426	& 29.3	& 37.5\\
 \hline
 White/Asian Applicants &  3.260 & 0.401 &  37.5	& 4.9\\

 \hline
 
\end{tabular}
\label{table:Counterfactual_Stats}
\end{table}

\newblock

The data was pre-processed via two methods. First, we emulated the procedure described in \citep{Counterfactual_Fairness}. The training attributes were therefore adjusted by treating the conditional expectations $\mathbb{E}(x|a_p=0)$ and $\mathbb{E}(x|a_p=1)$ as the respective contributions of the protected attribute to the training variables for White/Asian and Hispanic/Black applicants. The residuals $\tilde{x}(x,a_p) = x - \mathbb{E}(x|a_p)$ are then the new ``fair'' training variables.

\newblock

Subsequently, we adjusted the data using the counterfactual fairness technique described above. Technically, the training variables (both the LSAT scores and GPAs) are both discrete attributes, but they can take a broad range of values, and can therefore be subject to the same treatment as continuous attributes. For every possible discrete value taken by either of the training variables, $x_k$, we evaluate the value of both the conditional and joint cumulative distribution functions:
\begin{equation}
    P(x \leq x_k|a_p = a)=\frac{\sum_{i=1}^{N} I(x_i \leq x_k) I(a_p = a)}{\sum_{i=1}^{N}I(a_p = a)}.
\end{equation}
Here $I(b)$ is an indicator function equal to unity if $b$ is true and 0 otherwise. An analogous expression holds for the joint but now without the dependence on the protected attribute. All `percentile' values from both demographics are subsequently mapped to the value which has the same percentile in the joint distribution, such that:

\begin{equation}
    x_k \rightarrow{} \tilde{x_k} \hspace{2mm}\text{ where }\hspace{2mm} P(x \leq x_k|a_p) = P(x \leq \tilde{x_k}).
\end{equation}

\begin{figure}[h!]
\centering
\includegraphics[scale=0.4]{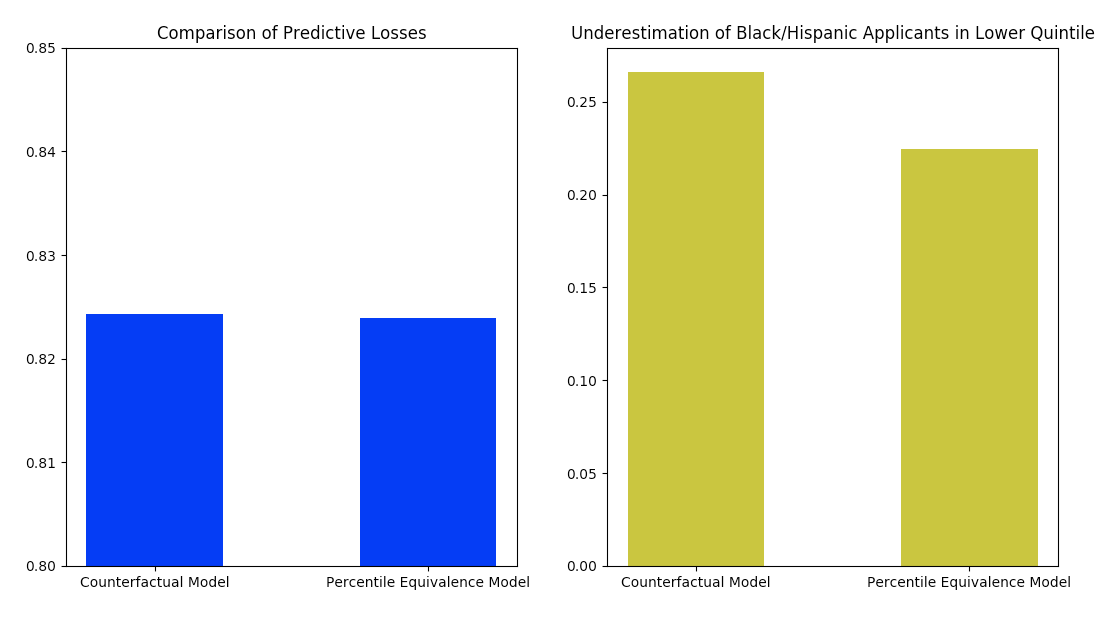}
\caption{Comparison of predictive Loss and Underestimation of black scores for the two models.}
\label{fig:Percentile_underestimation}
\end{figure}

Once this adjustment is complete, we can train a model on the adjusted attributes $\tilde{x_k}$. We trained a simple linear regression model on the two sets of adjusted data using a 5-fold cross validation procedure, using the square-error as the loss function:  $L(Y,\hat{Y}) = \sum_i(\hat{Y}_i - Y_i)^2$. Note that $\hat{Y}_i$ are the individual predictions of the model. The first year averages of candidates are the values to be predicted. The average performance of both models was found to be almost identical, as shown in figure \ref{fig:Percentile_underestimation}.

\newblock

Both models tended to slightly underestimate the score of black applicants, but the percentile equivalence pre-processing method performed better in this regard the data adjusted using a linear model. The difference in average underestimation of black/hispanic applicants in the bottom quartile of their distribution is shown in figure \ref{fig:Percentile_underestimation}.

\newblock

These findings are encouraging. They indicate that even in distributions where there is a small discrepancy in the standard deviation of the underlying conditional distributions, the percentile equivalence pre-processing technique can lead to minor improvements in the fair treatment of individuals at the tail end of distributions.

\newblock

After comparing the models on the validation set, we trained our percentile equivalence model on the test set and found a mean square error of $0.818$. It should be noted that this model achieved near perfect demographic parity on the test set, with an mean difference in predicted score between White/Asian applicants and Black/Hispanic applicants of just $\mathbf{-2.3 \times 10^{-4}}$. 

\section{Summary and Conclusions}


In this paper, we have demonstrated that SPNs can be a useful tool for implementing fair machine learning algorithms. We focused on a pre-processing strategy for implementing fairness, and demonstrated that the SPN is capable of identifying mutually independent subsets of variables in a multivariate data set. Those variables that belonged to a separate subset from the protected attribute were then regarded as `safe' variables which could be used to train a fair model, since no mutual information about the protected attribute is contained in these variables.

\newblock

We also consider how the remaining variables that exhibit protected attribute dependency can be adjusted to remove the protected attribute contribution. In the case of numerical variables, an additive model (equation \ref{eq:Adjust}) was constructed by performing conditional inference queries in the SPN. Some of the issues posed by qualitative valued variables are also discussed, and we proposed a simple model for removing the probabilistic contribution of the protected attribute from such categorical variables. However, we also highlight that this approach may be insufficient when dealing with complex models.

\newblock

Despite our concerns, we find that our scheme reduces disparate treatment of males and females across a variety of classifiers when applied to the German Credit data set \citep{German_Credit_Source}. We tested two models, one `pre-processing' technique where the protected attribute contributions are removed prior to training the model, and another `post-processing' technique, where the model is allowed to see the unadjusted data when training, but the protected attribute contributions are then removed at test time. There wasn't any significant discrepancy in classification accuracy between the two fair models; however, the pre-processing technique was generally found to lead to slightly lower levels of disparate treatment. The fair models generally tended to have slightly lower classification accuracies than the `fairness through unawareness' model, but generally only of the order of 1\%, which may represent an improvement on certain previous methods \citep{Learning_Fair_Reps}.

\newblock

We then consider a hypothetical situation in which the contribution of the protected attribute to the other training variables is not additive. We therefore develop the idea of percentile equivalence to adjust the distribution in such a way that $P(X|a_p) = P(X) \text{ } \forall \text{ } a_p$. We applied this technique to a data set containing information about the admission scores of college applicants \citep{lsac_information}. In general the percentile equivalence technique led to slightly better predictions for black/hispanic applicants at the bottom end of their own distribution, and therefore reduced the average underestimation of their scores.

Overall, this paper takes  first steps in how tractable models, especially tractable learning, can contribute to the research agenda of fairness in machine learning. Our initial results are very encouraging, and throughout the paper we motivated and discussed other definitions and concerns. In general, the ability of SPNs and other arithmetic circuit-based models to deal with large datasets (as evidenced in \cite{SPN_structure_learning,liang2017learning,molina2018mixed,bueff2018tractable,maxSPN}) will mean that probabilistic conditional queries can be computed exactly in service of robustly enabling 
fairness over big uncertain data. To that end, we hope to have provided key insights on how that is possible.

\bibliographystyle{abbrv}

\end{document}